\documentclass[lettersize,journal]{IEEEtran}
\usepackage{amsmath,amsfonts,bbding}
\usepackage{algorithmic}
\usepackage{algorithm}
\usepackage{array}
\usepackage{booktabs}
\usepackage{multirow}
\usepackage{tabularx} 
\usepackage[caption=false,font=normalsize,labelfont=sf,textfont=sf]{subfig}
\usepackage{textcomp}
\usepackage{stfloats}
\usepackage{url}
\usepackage{verbatim}
\usepackage{graphicx}
\usepackage{cite}
\usepackage{color,xcolor}
\usepackage{colortbl} 
\begin{document}

\title{X-Fake: Juggling Utility Evaluation and Explanation of Simulated SAR Images}


\author{Zhongling Huang,~\IEEEmembership{Member,~IEEE,}
Yihan Zhuang,
Zipei Zhong,
Feng Xu,~\IEEEmembership{Senior Member,~IEEE,}\\
Gong Cheng,
Junwei Han,~\IEEEmembership{Fellow,~IEEE}
\thanks{This work was supported by the National Natural Science Foundation of China under Grant 62101459.}
\thanks{Z. Huang, Y. Zhuang, Z. Zhong, G. Cheng, and J. Han are with the BRain and Artificial INtelligence Lab (BRAIN LAB), School of Automation, Northwestern Polytechnical University. F. Xu is with the Key Laboratory for Information Science of Electromagnetic Waves (MoE), Fudan University. Corresponding Author: Gong Cheng and Feng Xu (huangzhongling@nwpu.edu.cn, fengxu@fudan.edu.cn, gcheng@nwpu.edu.cn).}
}

\markboth{Journal of \LaTeX\ Class Files,~Vol.~14, No.~8, August~2021}%
{Shell \MakeLowercase{\textit{et al.}}: A Sample Article Using IEEEtran.cls for IEEE Journals}


\maketitle

\begin{abstract}
SAR image simulation has attracted much attention due to its great potential to supplement the scarce training data for deep learning algorithms. Consequently, evaluating the quality of the simulated SAR image is crucial for practical applications. The current literature primarily uses image quality assessment (IQA) techniques for evaluation that rely on human observers' perceptions. However, because of the unique imaging mechanism of SAR, these techniques may produce evaluation results that are not entirely valid. The distribution inconsistency between real and simulated data is the main obstacle that influences the utility of simulated SAR images. To this end, we propose a novel trustworthy utility evaluation framework with a counterfactual explanation for simulated SAR images for the first time, denoted as \textit{X-Fake}. It unifies a probabilistic evaluator and a causal explainer to achieve a trustworthy utility assessment. We construct the evaluator using a probabilistic Bayesian deep model to learn the posterior distribution, conditioned on real data. Quantitatively, the predicted uncertainty of simulated data can reflect the distribution discrepancy. We build the causal explainer with an introspective variational auto-encoder (IntroVAE) to generate high-resolution counterfactuals. The latent code of IntroVAE is finally optimized with evaluation indicators and prior information to generate the counterfactual explanation, thus revealing the inauthentic details of simulated data explicitly. The proposed framework is validated on four simulated SAR image datasets obtained from electromagnetic models and generative artificial intelligence approaches. The results demonstrate the proposed \textit{X-Fake} framework outperforms other IQA methods in terms of utility. Furthermore, the results illustrate that the generated counterfactual explanations are trustworthy, and can further improve the data utility in applications.
\end{abstract}


\section{Introduction}


Synthetic Aperture Radar (SAR) is an important remote sensing technology in many applications, owing to its all-day and all-weather imaging ability. At present, intelligent SAR image interpretation based on deep learning is facing the challenge of limited annotated data under abundant and varied imaging conditions \cite{huang2019and,wang2021sar,huang2017transfer}. Specifically, the characteristics of SAR targets vary dramatically with imaging parameters, especially the target orientation angle. It will have a significant impact on a deep learning model's generalization ability with limited measurements. Thus, generating high-quality SAR target images with various observation angles is of paramount importance for practical applications. Synthesizing SAR image data with electromagnetic models or advanced artificial intelligence technologies has attracted the most attention in recent years \cite{caoDemandDrivenSARTarget2022,wang2018synthetic,hu2021feature}. In this field, how to evaluate the quality of the simulated SAR images \cite{guo2022recognition,yu2022new} and how to utilize the simulated data properly to improve real-world applications are two important issues to be considered \cite{han2024improving,sun2023gradual,zhu2022lime}.

\begin{figure}[!tbp]
    \centering
    \includegraphics[width=0.5\textwidth]{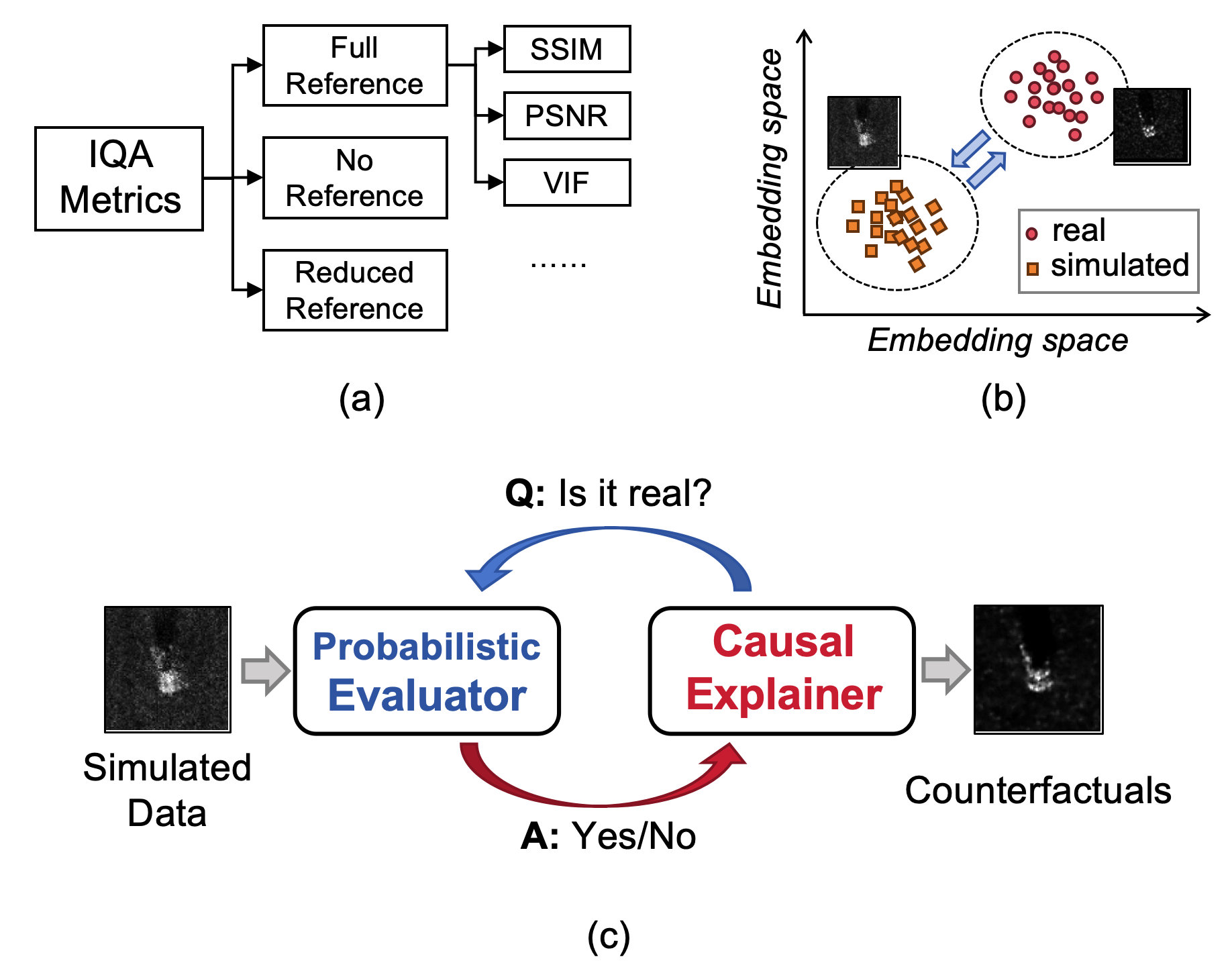}
    \caption{(a) Conventional IQA metrics for simulated SAR image evaluation. (b) The inconsistent distribution is the main obstacle that influences the utility of simulated data. (c) \textbf{Proposed \textit{X-Fake}:} A trustworthy framework that can evaluate and explain the simulated SAR images in terms of utility, with providing quantitative indicators as well as explicit high-quality explanations.}
    \label{fig:intro}
\end{figure}

The current research mainly tackles the above two issues separately. Fig. \ref{fig:intro} (a) illustrates the widespread application of image quality assessment (IQA) metrics for objective evaluation of simulated SAR data. The majority of related literature employs IQA metrics, like the structural similarity index measure (SSIM) and peak signal-to-noise ratio (PSNR) \cite{jiao2013sar}, for evaluation, a process that relies on human visual perception. However, due to the specific microwave characteristics of the SAR image, these metrics may not be applicable to SAR. Even though there are other SAR-specific image quality criteria, like equivalent look number (ELN) and radiometric resolution \cite{vespe2012sar}, they can only judge the SAR image based on how it looks and not how useful it is for building a deep model.


Many advanced simulation approaches can synthesize SAR images with good visual quality \cite{lewis2019sar}, but they have difficulties in real-world applications \cite{song2021learning,liu2018sar}. The inconsistent data distribution between simulated and real data would be one of the most significant obstacles that influences the utility, as presented in Fig. \ref{fig:intro} (b). The different physical parameters in model-based simulation, for example, would result in a background discrepancy between simulated and measured data. Researchers proposed both explicit and implicit approaches to address this issue. One of them is to use image pre-processing on simulated SAR data, like filtering and Gaussian noise \cite{choi2018despeckling,makitalo2010denoising}, which changes the simulated data itself. However, these strategies require manual and empirical design, which limits their flexibility. The implicit methods mainly include transfer learning and domain adaptation, aiming at learning representative knowledge from simulated data applicable to measured ones \cite{shi2022unsupervised}. However, they fail to provide a clear explanation for why the simulated SAR images lack utility. From the user's perspective, it is crucial to be able to figure out the flaws of generated data at the level of human understanding so as to improve the data utility and the simulation method.


To address the above issues simultaneously, we propose a novel unified framework, denoted as \textit{X-Fake}, to evaluate and explain the simulated SAR image in terms of utility. As shown in Fig. \ref{fig:intro} (c), \textit{X-Fake} comprises a probabilistic evaluator and a causal explainer collaborated together. The evaluator is able to answer the question: \textit{"Is the input real or not?"} with quantitative indicators. If the answer is "no", the causal explainer will generate the counterfactual image conditioned on the evaluation indicators and SAR target priors. We claim that the low utility of simulated SAR images, although they have good visual quality, lies in the significant distribution inconsistency. It may appear in inaccurate strong scattering points and target orientation given an azimuth angle, or distinct clutter distribution in the background. Consequently, we expect that the probabilistic evaluator can tell the quantified criteria that reflect the inconsistency, and the causal explainer can figure out the details of the simulated SAR image that lead to inauthenticity.

The proposed framework is inspired by Counterfactual Latent Uncertainty Explanations (CLUE), which was the first to explain the uncertainty estimated by the differentiable probabilistic model \cite{antoran2020getting}. The Bayesian neural network (BNN) takes the weight parameter as a probability distribution instead of a point estimation. It offers the opportunity to predict high uncertainty in the input when it is an out-of-distribution sample with training data. Therefore, we can consider the predicted uncertainty as a criterion that signifies the discrepancy in the distribution between real and simulated data. Considering the specific characteristics of the SAR target, we propose to construct a BNN model to estimate the class label and the azimuth angle simultaneously. In our work, we consider the combination of the predicted label and angle uncertainty to be a quantitative metric.

The goal of interpreting the predicted uncertainty is to answer the following question: \textit{"What would it be like if it could be considered a real SAR target?"} We observed that most simulated SAR images have good visual quality, and only some details restrict their utility. We introduce the counterfactual explanation (CE) to highlight the small changes to the input simulated data that would reduce the assigned uncertainty. Different from the previous CLUE method, we propose to construct the causal explainer with an IntroVAE model to ensure the high resolution of the generated counterfactual images. We also propose a multi-objective optimization strategy based on the simulated data and evaluation indicators. This allows us to clearly demonstrate the inauthentic details.



The contributions are summarized as follows:

\begin{itemize}
    \item A novel trustworthy utility evaluation framework with counterfactual explanation is proposed for simulated SAR images. To the best of our knowledge, the proposed \textit{X-Fake} offers the first attempt to simultaneously carry out the quantitative assessment and explain the inauthentic details of simulated data in terms of utility.
    
    
    
    \item In this framework, a Bayesian deep convolutional neural network (BDCNN) with category and angle prediction is constructed as the probabilistic evaluator. In regard to explaining the deficiency of simulated data, we propose an IntroVAE-based model optimized with multiple objectives based on evaluation indicators and SAR target priors to generate a high-resolution counterfactual image.
    
    
    \item Several simulated SAR image datasets are explored for validation, including AI-generated and electromagnetic model-simulated data. The proposed \textit{X-Fake} outperforms other IQA methods for global and individual evaluation. Additionally, the counterfactual explanations are intuitive, clear, and trustworthy, validated by extensive ablation studies, comparative studies, as well as quantitative and qualitative analysis.
\end{itemize}

Here is a summary of the paper's organization: Section \ref{sec:related} introduces the literature review related to this work regarding IQA methods, the utility of simulated SAR images, and uncertainty quantification with explanation. Then, Section \ref{sec:method} illustrates the proposed trustworthy utility evaluation and explanation framework \textit{X-Fake} for simulated SAR images. The experiments and result analysis are demonstrated in Section \ref{sec:exp}. Finally, Section \ref{sec:conclusion} presents the summary and outlooks, as well as the limitations of this work.

\section{Related Work}
\label{sec:related}

In this section, various works proposed for image quality assessment, evaluating and improving the utility of simulated SAR images, as well as uncertainty quantification and explanation, are briefly reviewed.

\subsection{Image Quality Assessment}

Image quality assessment (IQA) methods mainly aim to assign a score to an image that can evaluate its quality from a visual perspective. Subjective and objective assessment are the two main types of IQA methods. Subjective assessment measures quality with human input, while objective assessment measures quality without it \cite{mohammadi2014subjective,athar2019comprehensive}. The objective IQA, which is the most popular method, can be divided into full-reference (FR), no-reference (NR), and reduced-reference (RR) approaches \cite{wang2006modern}. A lot of research papers have used FR metrics like mean square error (MSE), peak signal-to-noise ratio (PSNR), structural similarity (SSIM), and visual information fidelity (VIF) to describe the quality of a simulated SAR image. These methods assess the image quality based on human visual perception. For generated images with deep learning models, the Fréchet Inception Distance (FID) is also widely used in feature distribution \cite{heusel2017gans}. In essence, it utilizes the pre-trained ImageNet Inception network as the feature extractor and compares the distances between features. However, due to the specific imaging mechanism of SAR, there may be a lack of trustworthiness for evaluation \cite{wang2018synthetic,wang2023improved}. The NR models rely primarily on image quality benchmark datasets that record human beings' evaluation results for visual perception \cite{mittal2012no,talebi2018nima}. However, the quality assessment of human visual systems is not applicable to SAR, as we will discuss in our later experiments.


\subsection{Utility of Simulated SAR Images}

In real-world applications, we mostly train the deep neural network with simulated SAR images and real data to enhance the algorithm's generalization ability. As a result, the utility of the simulated SAR images is worth studying. Some related work focuses on utility evaluation in terms of model performance. The SAR target recognition task, for example, is the most explored. For evaluation, Guo et al. proposed the Substitution Rate Curve (RSC) criteria based on the changing recognition rate \cite{guo2022recognition}. Similarly, the hybrid recognition rate curve (HRR) and the class-wise image quality assessment were proposed in the literature \cite{yu2022new}, respectively. Besides the utility evaluation, some other works aim to improve the data utility of simulated SAR images with transfer learning, domain adaptation, or knowledge distillation \cite{han2024improving,sun2023gradual,10217035}. These methods reduce the distribution discrepancy between real and simulated SAR images, enhancing model performance through the use of simulated data. In summary, evaluating and improving the utility of the simulated SAR images are equally important, yet current studies approach them separately. Mostly, the utility improvement takes place implicitly in the feature space, making it difficult to interpret. In contrast, our method aims to address the two issues simultaneously with a unified, trustworthy, and explainable framework.

\subsection{Uncertainty Quantification and Explanation for Trustworthy AI}

Methods for estimating uncertainty in a DNN prediction are a popular and vital field of research \cite{gawlikowski2023survey}. They can be categorized into single deterministic \cite{sensoy2018evidential}, Bayesian \cite{wilson_bayesian_2022}, ensemble \cite{sagi2018ensemble}, and test-time augmentation methods \cite{ayhan2022test}. For Bayesian neural networks, specifically, the posterior distribution over the network parameters is estimated, and the predictive uncertainty of test data can be quantified. Approaches such as variational inference \cite{blundell2015weight}, Monto Carlo sampling \cite{gustafsson2020evaluating}, and Laplace approximation \cite{ritter2018scalable} were proposed to infer the posterior distribution of BayesCNN. Apart from uncertainty estimation, it is also important to explain the source of uncertainty in order to achieve more trustworthy artificial intelligence (AI). Researchers in explainable artificial intelligence (XAI) have developed numerous methods for helping users understand the predictions of complex supervised learning models \cite{samek2021explaining}. By contrast, explaining the uncertainty of model outputs has received relatively little attention \cite{watson2024explaining,thiagarajan2021explanation,huang2023uncertainty}. The mainstream explanation methods can tell why a model works well, but they will be ineffective in explaining the bad result with very high uncertainty \cite{8237336}. Some researchers conducted causal counterfactuals to explain the high uncertainty in the input \cite{schwab2019cxplain}. Recently, the literature \cite{antoran2020getting,ley2022diverse} proposed methods that can help identify which input features are explicitly responsible for models' uncertainty.

Inspired by the introduced work, we re-formulate the utility evaluation problem by quantifying the domain-shift uncertainty of simulated SAR images in this paper. Furthermore, we illustrate the inauthentic details of the fake SAR image through an uncertainty explanation.

\section{Method}
\label{sec:method}

\subsection{Overview}

\begin{figure*}[!htbp]
    \centering
    \includegraphics[width=0.7\textwidth]{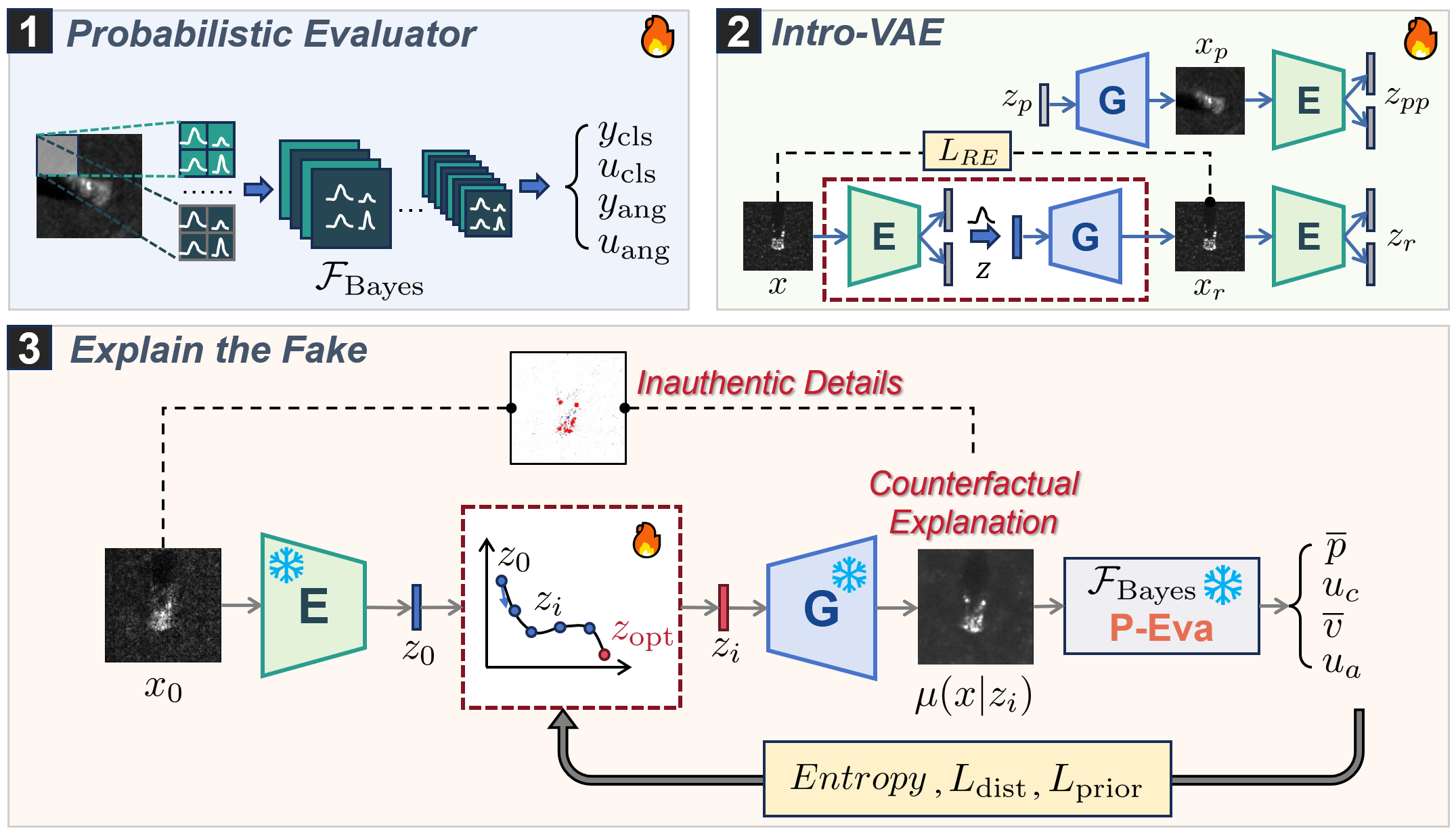}
    \caption{The proposed trustworthy utility assessment framework, \textbf{X-Fake}. (1) Constructing a probabilistic evaluator (P-Eva) with Bayesian deep convolutional neural networks, consisting of label and azimuth angle prediction. (2) Pre-training the IntroVAE model to prepare for high-quality counterfactual generation. (3) Optimizing the latent code to obtain the counterfactual explanation. }
    \label{fig:method}
\end{figure*}

The proposed \textit{X-Fake} framework, as shown in Fig. \ref{fig:method}, consists of three steps. The probabilistic evaluator (P-Eva) is firstly constructed with a BayesCNN model. The simulated SAR image $x_0$ is evaluated by P-Eva to obtain the criteria vector $\mathbf{m}$ containing target class and angle predictions and their uncertainties, denoted as:
\begin{equation}
    \mathbf{m}_0 = \mathcal{F}_{\mathrm{Bayes}}(x_0) = [\overline{y}, u_c, \overline{v}, u_a],
\end{equation}
where $\mathcal{F}_{\mathrm{Bayes}}(\cdot)$ represents the BayesCNN model. The construction and optimization details will be introduced in Section \ref{sec:probeva}. Then, an IntroVAE model is pre-trained to obtain the optimized encoder $\mu_{\phi}$ and generator $\mu_{\theta}$, thus preparing for high-quality counterfactual generation. It can be denoted as:
\begin{equation}
    \theta_{G},\phi_{E}=\mathop{\arg\min}\limits_{\theta,\phi}L(\theta,\phi;x).
\end{equation}
The details will be elaborated in Section \ref{sec:introvae}. Given the simulated SAR image $x_0$, its counterfactual explanation is defined as $x_c$ that satisfies the following conditions. On the one hand, $x_c$ should be close to $x_0$, where the distance metric $d(\cdot,\cdot)$ is applied for constraint. On the other hand, $x_c$ should be satisfied with the desired output $\mathbf{m}_c$ of P-Eva. To this end, the optimization can be written as:
\begin{equation}
    x_c=\mathop{\arg\max}\limits_{x}(p(\mathbf{m}_c|x)-d(x,x_0)).
\end{equation}
Note that optimizing $x_c$ in the high-dimension image space is difficult. To this end, we transform the optimization problem in the latent space of the pre-trained IntroVAE model. The algorithms will be illustrated in details in Section \ref{sec:counterfactual}.



\subsection{Probabilistic Evaluator}
\label{sec:probeva}

The current SAR target image generation is based on prior parameters, and among them the target class and azimuth angle are most critical. The inauthentic details of a simulated SAR target image exist in the following three aspects: ambiguous scattering centers, inaccurate azimuth angles, and out-of-distribution clutter background. They can be summarized with inconsistent feature distribution between real and simulated SAR images, which leads to inferior utility of simulated data. To this end, we propose to construct a probabilistic evaluator based on Bayesian deep neural network to measure the feature distribution discrepancy in terms of class label and azimuth angle.
\begin{figure}[H]
    \centering
    \includegraphics[width=0.3\textwidth]{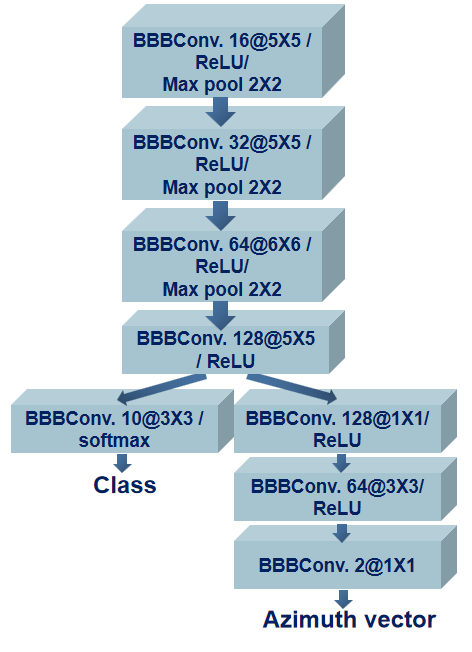}
    \caption{ Overall architecture of the proposed BBB-A-ConvNet-a. The BBB convolution layers are represented as “BBBConv. (number of feature maps) $@$ (filter size).”}
    \label{fig:BBB_a}
\end{figure}
Specifically, we construct a BayesCNN model as shown in Fig. \ref{fig:BBB_a}. It features four trainable layers in the front and is subsequently bifurcated into two branches, one for target category prediction and the other for azimuth vector regression prediction. The azimuth $\theta$ is encoded as a vector $[\cos\theta, \sin\theta]$. In this manner, our BayesCNN network is capable of simultaneously predicting the category label and azimuth angle of the SAR target image.

Compared to its equivalent frequencist CNN, which utilizes a single point estimate for weight parameters, BayesCNN incorporates a probability distribution for its weights. This approach involves conducting posterior inference on the distribution parameters of the neural network using Bayes' theorem. Therefore, BayesCNN offers results by weighting the posterior probability of all parameter settings, rather than relying solely on a deterministic value of parameters. This approach allows for not only predicting results, but also capturing the uncertainty of those predictions. Our method involves assessing the utility of simulated data by quantifying the uncertainty of the recognition model outputs.

The BayesCNN construction and optimization can be realized in different ways. In this paper, we applied optimization with Bayes by Backprop and Monte Carlo Dropout approximation to construct the P-Eva, denoted as Eva-BBB and Eva-MCD, respectively.


\subsubsection{Optimization with Bayes by Backprop}

Backpropagation and gradient descent are widely applied for neural network training. We apply one of the most popular BayesCNN optimization approaches, Bayes by Backprop (BBB) \cite{shridhar2019comprehensive} to learn the posterior distribution of BayesCNN parameters. BBB is a variational inference-based method to learn the posterior distribution on the weights of a neural network from which weights $\omega\sim q_\theta(\omega|D)$ can be sampled in back propagation. It regularizes the weights by maximizing the Evidence Lower Bound (ELBO).

To learn the Gaussian distribution parameters that weight follows, we need to find a way to approximate the intractable true posterior distribution $p(\omega|D)$. We use the method of variational inference. This is done by minimizing the Kullback-Leibler difference between the simple variational distribution $q_\theta(\omega|D)$ and the  true posterior distribution $p(\omega|D)$, that is,
\begin{equation}
\begin{aligned}
& \mathop{\arg\min}\limits_\theta KL[q_\theta(\omega|D)||p(\omega|D)] \\
\end{aligned}
\end{equation}
It can be achieved by maximizing the ELBO, that is:
\begin{equation} 
\begin{aligned}
& \mathop{\arg\min}\limits_\theta -ELBO \\
= & \mathop{\arg\min}\limits_\theta E_{q_\theta}[\log (p(D|\omega))]-KL[q_\theta(\omega|D)||p(\omega)].
\end{aligned}
\end{equation}
We use the Monte Carlo method to obtain the final objective function approximately, that is,
\begin{equation}
\label{equ:loss}
\mathop{\arg\min}\limits_\theta \sum_{i=1}^{n} \log q_\theta(\omega^{(i)}|D) - \log p(\omega^{(i)})- \log p(D|\omega^{(i)})
\end{equation}
where $n$ is the number of draws.

Note that the last term in Equation (\ref{equ:loss}) refers to the maximum likelihood objective, which is often realized by cross-entropy (classification) or mean square error (regression). As a result, $\log p(D|\omega^{(i)})$ can be written as:

\begin{equation}
\label{equ:evaloss}
    \log p(D|\omega^{(i)}) = \sum_{j=1}^C \hat{y} \log p(y_{i,j}|x) - \lambda_a ||v_i-\hat{v}||_2^2,
\end{equation}
where $\lambda_a$, $\hat{y}$, $\hat{v}$ are hyper-parameter, ground truth of class label and angle, respectively. 

The local reparameterization trick is applied for optimizing. According to Kumar Shridhar's work \cite{shridhar2019comprehensive}, the reparameterization in convolutional layer is achieved by two convolutional operations. Denote the random variable $\omega_{i}$ as convolutional filters in the $i$th layer, and the variational posterior probability distribution as $q_{\theta}(\omega_{i}|D)=N(\mu_{i},\alpha_{i}\mu^{2}_{i})$. The output of the $i$th layer can be reparameterized as:
\begin{equation}
\label{equ:reparam}
\mathbf{A}_{i+1} = \mathbf{A}_i * \mu_i + \epsilon \odot \sqrt{\mathbf{A}_{i}^2*(\alpha_i\odot\mu_{i}^2)},
\end{equation}
where $\epsilon \sim N(0,1)$. $\mathbf{A}_i$ is the input feature map of the $i$th layer. $*$ and $\odot$ denote the convolution operation and component-wise multiplication, respectively. 

Thus, the output of a Bayesian convolution layer can be realized by two steps of convolutions, that is, a convolution with the input feature map $\mathbf{A}_i$ and the mean values of kernels and another convolution with the square of $\mathbf{A}_i$ and the variances of kernels. In this way, the parameters $\mu_i$ and $\alpha_i$ can be updated separately in the two steps of convolution.

\subsubsection{Monte Carlo Dropout as a Bayesian Approximation}

Gal et al. \cite{gal2016dropout} have demonstrated that a neural network with arbitrary depth and non-linearities applied before each weight layer is mathematically equivalent to an approximation of a probabilistic deep Gaussian process. According to this method, we incorporate dropout layers preceding each convolutional layer. Unlike its equivalent frequencist CNN, dropout is applied not only during model training but also during testing in order to approximate a probabilistic deep Gaussian process. The BayesCNN constructed using this method is named Eva-MCD.

\subsubsection{Uncertainty}

The category uncertainty of BayesCNN prediction $u_c$ is given by \cite{shridhar2019comprehensive}:
\begin{equation} 
u_c=\underbrace{\frac{1}{T}\sum_{t=1}^{T}diag(\hat{y}_t)-\hat{y}_t\hat{y}_t^T}_{aleatoric}
+\underbrace{\frac{1}{T}\sum_{t=1}^{T}(\hat{y}_t-\overline{y})(\hat{y}_t-\overline{y})^T}_{epistemic},
\end{equation}
where $\hat{y}_t=Softmax(f_{\omega_t}(x))$ denotes the frequentist inference with parameter $w_t$ sampled from the obtained posterier distribution, and $\overline{y}=\frac{1}{T}\sum_{t=1}^{T}\hat{y}_t$ denotes the Bayesian average of predictions, and $T$ is the sampling number.

The uncertainty for azimuth angle prediction $u_a$ can be expressed as the variance of multiple samplings: 
\begin{equation} 
u_a=\frac{1}{T}\sum_{t=1}^T(v_t-\overline{v})^2,
\end{equation}
where $\overline{v}=\frac{1}{T}\sum_{t=1}^{T}v_t$.

To sum up, the output of P-Eva $\mathbf{m}$ containing category and azimuth angle predictions with the corresponding uncertainties can be given by:
\begin{equation} 
\mathbf{m} = [\overline{y}, u_c, \overline{v}, u_a]
\end{equation}
The overall uncertainty can be calculated as $u_c+u_a$.

\subsection{IntroVAE Pre-training}
\label{sec:introvae}

Optimizing the counterfactual images in the high-dimensional space is challenging and cannot ensure the in-distribution constraint \cite{antoran2020getting}. To this end, the optimization can be conducted in the latent space of a probabilistic generative model to ensure the similar data manifold. In order to generate precise counterfactual explanations, it is crucial to first develop a high-quality image generation model. Consequently, we propose to utilize IntroVAE \cite{huang2018introvae} as the foundational structure of the explainer. Similar to VAE, IntroVAE consists of an encoder and decoder that can project the high-dimensional data into low-dimensional latent space, enabling us to optimize in the latent space in order to preserve the data manifold. Additionally, IntroVAE incorporates adversarial training based on VAE to enhance the quality of the generated counterfactual explanations.

As depicted in Fig. \ref{fig:method}, the VAE encoder is additionally served as the discriminator, and the VAE decoder is served as the generator in GAN. Adversarial learning like GAN is performed during model training so that the model can self-estimate the difference between the generated sample and the training data, and then update itself to produce a more realistic sample. 

In order to match the distribution of the generated samples to the true distribution of the given training data, we use the regularization term $L_{\mathrm{KL}}$ as the adversarial training cost function. The Encoder requires minimizing $L_{\mathrm{KL}}(z)$ so that the posterior $q_{\phi}(z|x)$ of the real data $x$ matches the prior $p(z)$. Simultaneously, it requires maximizing $L_{\mathrm{KL}}(z_s)$ so that the posterior $q_{\phi}(z_s|x_s)$ of the generated sample deviates from the prior $p(z)$. $z_s$ includes $z_r$ and $z_{pp}$ which are the latent codes of the reconstructed images $x_r$ and $x_p$, respectively.

The loss function of the Encoder is:
\begin{equation} 
L_E = \alpha_R L_{\mathrm{KL}}(z)+\alpha_E\sum_{s=r,pp}[m-L_{\mathrm{KL}}(z_s)]^++\beta L_{\mathrm{RE}}(x,x_r)
\end{equation}

Conversely, the Generator needs to minimize $L_{\mathrm{KL}}(z_s)$ so that the posterior $q_{\phi}(z_s|x_s)$ approximately matches the prior $p(z)$:
\begin{equation} 
L_G = \alpha_G\sum_{s=r,pp}L_{\mathrm{KL}}(z_s)+\beta L_{\mathrm{RE}}(x,x_r)
\end{equation}
where $m$ is a positive margin, $[.]^+=max(0,.)$. $L_{\mathrm{RE}}$ is the mean square error (MSE) loss function of the reconstruction error. $L_{\mathrm{KL}}$ represents KL-divergence: $L_{\mathrm{KL}}(z;\mu,\sigma)=\frac{1}{2}\sum_{i=1}^N\sum_{j=1}^{M_z}(1+log(\sigma_{ij}^2)-\mu_{ij}^2-\sigma_{ij}^2)$. $\alpha_R$, $\alpha_E$, $\alpha_G$ and $\beta$ are weighting parameters used to balance the importance of each item. 

\subsection{Counterfactual Explanation Generation}
\label{sec:counterfactual}

\begin{figure*}[!hbp]
    \centering
    \includegraphics[width=0.8\textwidth]{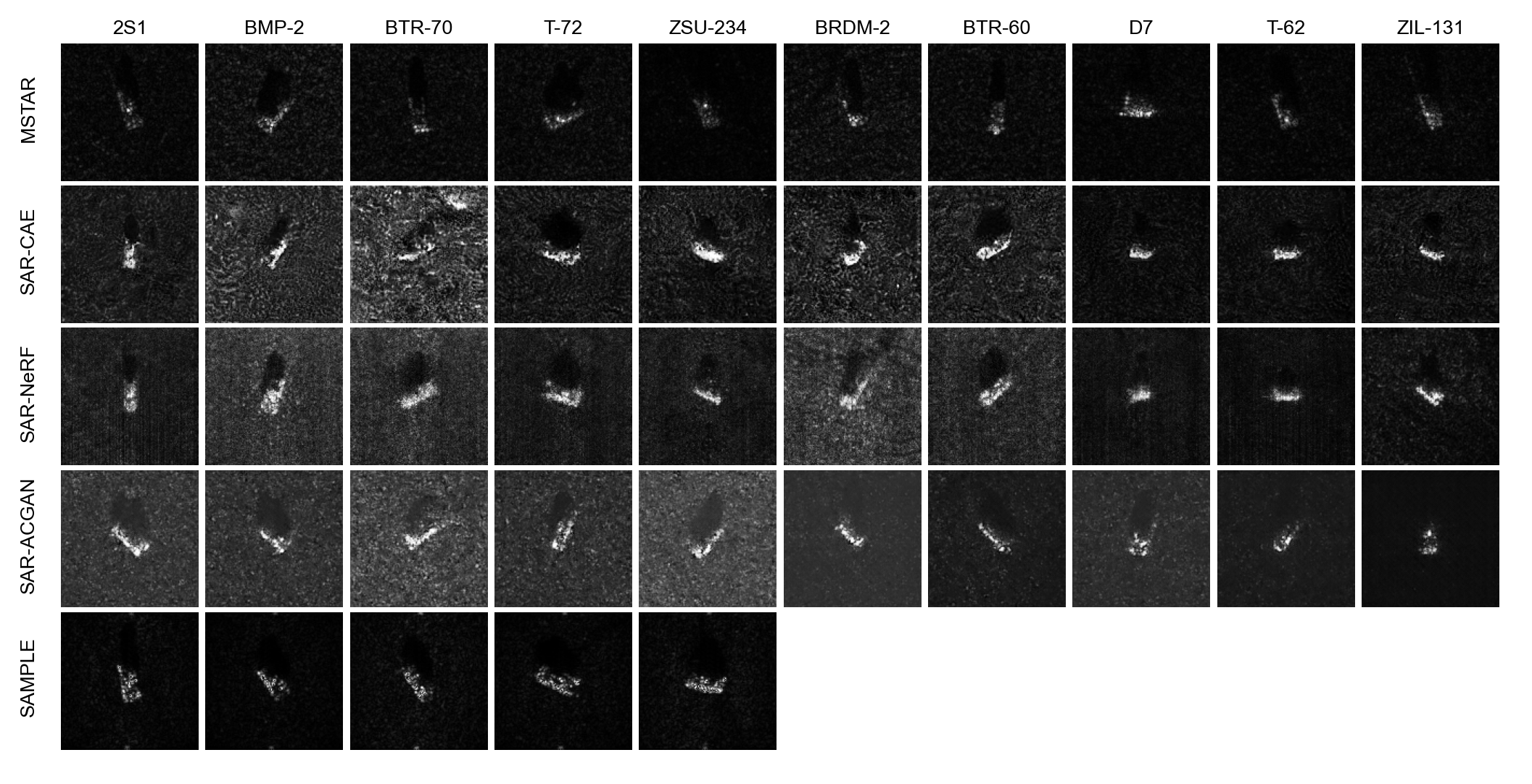}
    \caption{Some examples of the experimented datasets.}
    \label{fig:dataset}
\end{figure*}

Counterfactual analysis aims to examine the question of \textit{"What changes are required to the input in order to get the desired output?"} to explain how the model makes decisions. In our study, we aim to minimize the modifications required for SAR images deemed of low quality by the BayesCNN evaluation network in order to optimize their utility for deep learning models. This involves generating an "explanation" closely resembling the original simulated images, and enabling the evaluator to achieve the desired outcome.

For the input simulated SAR image $x_0$, the prediction of P-Eva is denoted as $\mathbf{m}_0 = [\overline{y}_0, u_c, \overline{v}_0, u_a]$ where the overall uncertainty $u_c+u_a$ could be relatively high, or the label and angle prediction $\overline{y}_0,\overline{v}_0$ are even wrong. The counterfactual explanation $x_{\mathrm{opt}}$ should meet the requirements of correct prediction $\overline{y}_{\mathrm{opt}},\overline{v}_{\mathrm{opt}}$, lower uncertainty, and minimal change compared to $x_0$. Therefore, the optimization can be divided into the following three parts. First is the distance constraint, denoted as:
\begin{equation}
    L_{\mathrm{dist}}(x) = d_1(x,x_0),
\end{equation}
where $d_1(\cdot,\cdot)$ is the L1 norm. The following constraints the counterfactual image consistent with the prior category and angle information:
\begin{equation}
    L_{\mathrm{prior}}(x) = -\lambda_y\sum_{i=0}^C \overline{y}_{\mathrm{opt}} log p(\overline{y}_i|x) + \lambda_v d_2(\overline{v}_{\mathrm{opt}},\overline{v}),
\end{equation}
where $C$ denotes the class number, and $d_2$ denotes the L2 norm. $\lambda_y, \lambda_v$ are two hyper-parameters. The final term defines the predicted uncertainty of $x$:
\begin{equation}
    L_{\mathrm{Entropy}}(x) = u_c + u_a
\end{equation}
The above-mentioned $\overline{y},\overline{v},u_c,u_a$ are predicted by the pre-trained P-Eva model. As a result, the overall optimization can be written as:
\begin{equation}
\begin{aligned}
\label{equ:L1}
L(x) & = L_{\mathrm{Entropy}}(x) + L_{\mathrm{prior}}(x) + \lambda_d L_{\mathrm{dist}}(x)
\end{aligned}
\end{equation}

In the optimization process, we do not directly optimize $x$ in the high-dimensional space, but map $x$ to the low-dimensional latent space through the pre-trained encoder of IntroVAE to optimize $z$.
Therefore, Equation (\ref{equ:L1}) can be re-written as: 
\begin{equation}
\begin{aligned}
\label{equ:L2}
L(z) = L_{\mathrm{Entropy}}(\mu_{\theta}(x|z)) & + L_{\mathrm{prior}}(\mu_{\theta}(x|z)) \\
& + \lambda_d L_{\mathrm{dist}}(\mu_{\theta}(x|z))
\end{aligned}
\end{equation}
Then, the latent counterfactual $z_{\mathrm{opt}}$ can be obtained from:
\begin{equation}
\label{equ:zopt}
    z_{\mathrm{opt}} = \mathop{\arg\min}\limits_z L(z).
\end{equation}
The counterfactual explanation in the image domain can be generated from the pre-trained decoder of IntroVAE, i.e., $x_{\mathrm{opt}}=\mu_{\theta}(x|z_{\mathrm{opt}})$. During the counterfactual generation, only the latent code $z$ is optimized while the P-Eva, the encoder and generator of IntroVAE are frozen.

The inauthentic details of a simulated SAR image can be reported as the pixel changes of the counterfactual explanations and the original image. For a better visualization, the difference $\Delta x = x_{\mathrm{opt}} - x_0$ is multiplied with its absolute value $|\Delta x|$ while keeping the sign. The positive and negative values are colored with red and blue in the experiments, respectively.

\section{Experiments}
\label{sec:exp}

\subsection{Datasets}

In the experiments, a real SAR image dataset and four simulated SAR image datasets are applied to conduct experiments. The simulated datasets include an EM model based simulation dataset and three generated SAR image datasets by advanced generative AI approaches. Some examples are given in Fig. \ref{fig:dataset}.


\begin{table*}[!htbp]
\centering
\caption{The experimented datasets include a real SAR image dataset (MSTAR) and four simulated SAR image datasets (SAR-ACGAN, SAR-CAE, SAR-NeRF, and SAMPLE). For AI-generated SAR images, the training numbers of generation are given in \textbf{Gen-Train}.}
\label{tab:dataset}
\resizebox{\linewidth}{!}{
\begin{tabular}{ccccccccccccccc}
\toprule
\multicolumn{2}{c}{Dataset}                         & Dep Angle & 2S1 & BMP-2 & BRDM-2 & BTR-60 & BTR-70 & D7  & T-62 & T-72 & ZIL-131 & ZSU-234 & \textbf{Total} & \textbf{Gen-Train} \\
\midrule
\multirow{2}{*}{Real}      & \multirow{2}{*}{MSTAR} & 15°       & 274 & 195   & 274    & 195    & 196    & 274 & 273  & 196  & 274     & 274     & 2425  & /  \\
                          &                        & 17°       & 299 & 233   & 298    & 256    & 233    & 299 & 299  & 232  & 299     & 299     & 2747 & /  \\
\midrule
\multirow{4}{*}{Simulated} & SAR-ACGAN               & 17°       & 299 & 233   & 298    & 256    & 233    & 299 & 299  & 232  & 299     & 299     & 2747 & 237 \\
                           & SAR-CAE                    & 17°       & 286 & 220   & 286    & 244    & 221    & 287 & 287  & 220  & 287     & 287     & 2625 & 121  \\
                           & SAR-NeRF               & 17°       & 360 & 360   & 360    & 360    & 360    & 360 & 360  & 360  & 360     & 360     & 3600 & 258 \\
                           
                           & SAMPLE                 & 15°-17°   & 174 & 107   & 0    & 0     & 92    & 0   & 0    & 108    & 0       & 174       & 655 & /  \\
\bottomrule
\end{tabular}}
\end{table*}

\subsubsection{MSTAR}

The Moving and Stationary Target Acquisition and Recognition (MSTAR) dataset \cite{mstar} is acquired by an X-band high-resolution synthetic aperture radar at Spotlight mode with a resolution of 0.3m $\times$ 0.3m. The dataset primarily comprise SAR slice images of stationary vehicles, encompassing target images of various vehicle types obtained at different azimuth angles. The experiments utilized data acquired at elevation angles of 15$^\circ$ and 17$^\circ$, containing images of 10 different classes, that is, 2S1, BMP-2, BRDM-2, BTR-60, BTR-70, D7, T-62, T-72, ZIL-131, and ZSU-234. The imaging azimuth angle ranges from 1$^\circ$ to 360$^\circ$, with the interval of 1$^\circ$ to 15$^\circ$.

\subsubsection{SAR-ACGAN}

The SAR-ACGAN dataset was generated by the Auxiliary Classifier Generative Adversarial Network (ACGAN) model \cite{odena2017conditional}, which was trained on 237 samples of MSTAR dataset at the depression angle of 17$^\circ$. The selected 237 training images were sampled with the azimuth angle interval of 15$^\circ$. The trained ACGAN model generated 2747 SAR images in total, as illustrated in Table \ref{tab:dataset}.

\subsubsection{SAR-CAE}

The SAR-CAE dataset is constructed with a causal adversarial autoencoder (CAE) model proposed by Guo et al. \cite{10309941}. CAE is a causal model for SAR image representation conditioned on disentangled semantic factors. It was trained on limited MSTAR data with a few azimuth angles, then it can generate SAR target images given an arbitrary azimuth angle and class label. Only 12 or 13 images per category were sampled from MSTAR dataset at the depression angle of 17$^\circ$ for training the CAE model, with an azimuth interval of 30$^\circ$. The number of the generated images is 2625, as given in Table \ref{tab:dataset}.


\subsubsection{SAR-NeRF}

SAR-NeRF is a cutting-edge generative AI model considering the imaging mechanism of SAR inspired by the concept of neural radiance fields (NeRF), proposed by Lei et al. \cite{lei2023sar}. The SAR-NeRF dataset is derived from the MSTAR dataset, with an azimuth interval of 15$^\circ$ and a depression angle of 17$^\circ$. The imaging azimuth angle interval of SAR-NeRF dataset is fixed at 1$^\circ$, resulting in a total of 3600 images with a depression angle of 17$^\circ$, as depicted in the third row of Table \ref{tab:dataset}.


\subsubsection{SAMPLE}

The Synthetic and Measured Paired and Labeled Experiment (SAMPLE) dataset \cite{lewis2019sar} provides EM model based simulation data and the paired measured SAR images with the same configuration as MSTAR dataset. In this paper, we selected the EM-simulated SAR images of same categories with MSTAR dataset for evaluation, including 2S1, BMP-2, BTR-70, T-72, and ZSU-234. The sampling azimuth angle ranges from 280$^\circ$ to 350$^\circ$ (10$^\circ$ to 80$^\circ$ according to the angle definition of MSTAR dataset), with elevation angles ranging from 15$^\circ$ to 17$^\circ$. The imaging interval is mainly about 1-2$^\circ$, but there are also some images with intervals ranging from 3$^\circ$ to 10$^\circ$. The details are given in Table \ref{tab:dataset}.

\subsection{Experimental Settings}
\begin{table}[!htbp]
\centering
\caption{Training set and validation set to train different simulated data evaluation networks and IntroVAE.}
\label{tab:evaluation dataset}
\begin{tabular}{ccccc}
\toprule
\textbf{}  & SAR-NeRF & SAR-CAE  & SAR-ACGAN & SAMPLE\\
\midrule
Training      & 15$^\circ$(80\%) & 15$^\circ$(80\%) & 15$^\circ$(80\%)  & 17$^\circ$(80\%) \\
Validation & 15$^\circ$(20\%) & 15$^\circ$(20\%) & 15$^\circ$(20\%)  & 17$^\circ$(20\%)\\
\bottomrule
\end{tabular}
\end{table}
In this paper, all experiments are conducted on NVIDIA GeForce RTX 3090 Ti with a compute capability of 8.6. The details of training and validation data settings are given in Table \ref{tab:evaluation dataset}. 

The proposed evaluation method aims to assess the distribution inconsistency between real and simulated data. Consequently, different from other full-reference assessment methods, our method does not require paired data. In order to prove the generalization ability of the proposed evaluation method, we applied real SAR images with different depression angle to train the BayesCNN model. To evaluate SAR-ACGAN, SAR-CAE and SAR-NeRF data with depression angle of 17$^\circ$, the BayesCNN model is trained with MSTAR real data at 15$^\circ$, as given in Table \ref{tab:evaluation dataset}.

The input size of evaluation model is 88$\times$88. The training data are pre-processed with center-crop, logarithm and random grayscale stretching for data augmentation, aiming to improve the generalization of the evaluation model. We used Adam as the optimizer with a learning rate of 0.001. The epoch number and the batch size are set to 300, 25, respectively. The trade-off parameter $\lambda_a$ in Equation (\ref{equ:evaloss}) is set to 20.

At IntroVAE Pre-training stage, the training and validation settings as well as the data pre-processing, follow the settings of evaluation stage. The hyper-parameters are given in Table \ref{tab:hyper-param-IntroVAE}. We use Adam as optimizer to optimize Generator and Encoder with a learning rate of 0.0005. The epoch is 500 and the batch size is 25.

\begin{table}[!htbp]
\centering
\caption{The hyper-parameters required to train IntroVAE.}
\label{tab:hyper-param-IntroVAE}
\begin{tabular}{cccccc}
\toprule
$\beta$ & $\alpha_R$ &$m$& $\alpha_E$  & $\alpha_G$ & latent dim \\
\midrule
10  & 0.0005 & 100 & 0.0005 & 0.0005 & 100 \\
\bottomrule
\end{tabular}
\end{table}

The trade-off parameters $\lambda_d$, $\lambda_y$, and $\lambda_v$ during the counterfactual optimization will be discussed in the experiments. The optimal values in our experiments are set to 1, 1, and 30, respectively. The Adam optimizer is applied, and the learning rate is 0.0005. The epoch numbers are set to 50, 150, 50, and 200 for SAR-NeRF, SAR-CAE, SAR-ACGAN, and SAMPLE, respectively.

\subsection{Quantitative Results of Evaluation}

In order to prove the effectiveness of the proposed evaluation method, we sorted the simulated data with different assessment metrics and split them into "TOP" and "LAST" evenly. Then, the classification model is trained with "TOP" and "LAST" data separately and tested with real images. The performance gap between "TOP" and "LAST" demonstrates the effectiveness of the assessment metrics in terms of utility.

\begin{figure*}[!htbp]
    \centering
    \includegraphics[width=1.0\textwidth]{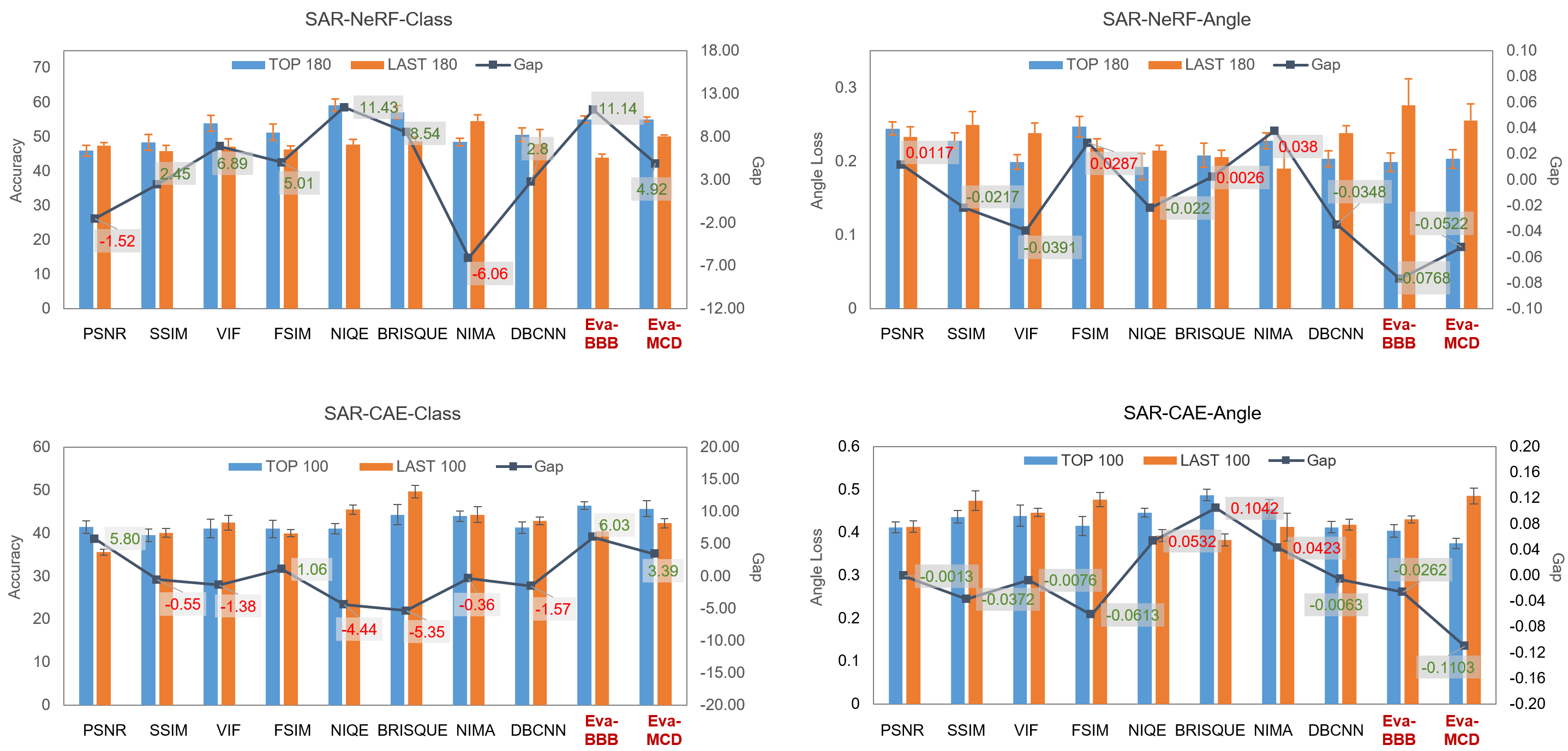}
    \caption{We compare four FR-IQA metrics (PSNR \cite{korhonen2012peak}, SSIM \cite{wang2004image}, VIF \cite{sheikh2006image}, FSIM \cite{5705575}) and four NR-IQA models (NIQE \cite{6353522}, BRISQUE \cite{6272356}, NIMA \cite{8352823}, DBCNN \cite{8576582}) with the proposed Eva-BBB and Eva-MCD. For each simulated SAR image dataset, we sort the samples according to the evaluation metrics and split them into "TOP" and "LAST" evenly. The deep models are trained with "TOP" and "LAST" data, respectively, with classification accuracy and angle loss reported in the results. "Gap" denotes the performance gap between "TOP" and "LAST", where the \textcolor{green}{green} values indicate the effectiveness of the evaluation method in terms of utility.}
    \label{fig:evaluation}
\end{figure*}

In this part, we applied SAR-NeRF and SAR-CAE datasets for experiments. The "TOP" and "LAST" numbers of SAR-NeRF and SAR-CAE are set to 180 and 100 per category, respectively. Four full-reference IQA metrics, i.e., PSNR \cite{korhonen2012peak}, SSIM \cite{wang2004image}, VIF \cite{sheikh2006image}, FSIM \cite{5705575}, and four non-reference IQA models, i.e., NIQE \cite{6353522}, BRISQUE \cite{6272356}, NIMA \cite{8352823}, DBCNN \cite{8576582}, are compared. 

As shown in Fig. \ref{fig:evaluation}, the classification accuracy of models trained with "TOP" data filtered by different assessment metrics are marked in blue, while the ones trained with "LAST" are presented in orange. The performance gap between "TOP" and "LAST" is also plotted. The positive values denote the assessment metrics are effective in terms of utility, otherwise demonstrating ineffective.

The results show that our method achieves the highest performance gap, which demonstrates that our method outperform other full-reference and non-reference IQA methods in terms of utility. Notably, some IQA models fail to evaluate the utility of simulated SAR image with negative performance gaps. The results illustrate that those methods based on human visual perception may not applicable to SAR images. We also explore the effectiveness of different BayesCNN models, including Eva-BBB and Eva-MCD. The results show that Eva-BBB can achieve better performance in evaluation.

\begin{figure}[!tbp]
    \centering
    \includegraphics[width=0.45\textwidth]{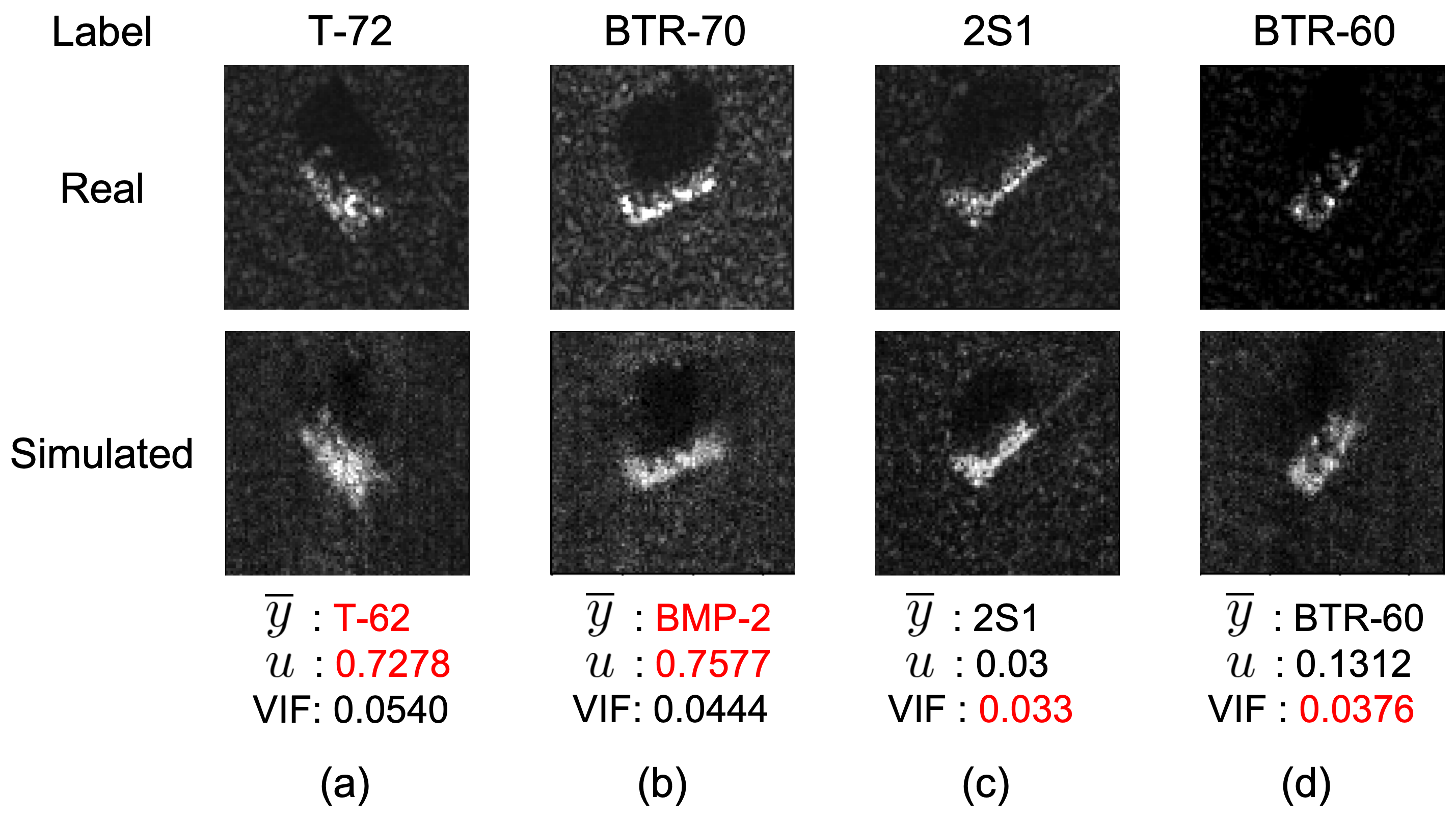}
    \caption{Some selected real and simulated images pairs are evaluated with VIF and Eva-BBB. The ground-truth and the predicted labels of simulated SAR images by a recognition model trained with real data are presented.}
    \label{fig:evaluation sample}
\end{figure}

Fig. \ref{fig:evaluation sample} exhibits four paired examples of real and simulated images. It is difficult to distinguish the quality of these four simulated images from visual perspective. The VIF indicators demonstrate they have similar image quality, and example (c) and (d) are with lower quality scores than (a) and (b). However, we observed the uncertainties of simulated sample (a) and (b) are high, denoting their distinct feature distribution discrepancy with real data. The incorrect predicted labels by a recognition model trained with real data can verify this. As a result, our method can successfully assess the simulated SAR images in terms of data utility, even they have similar visual quality.

\begin{figure*}[!hbp]
    \centering
    \includegraphics[width=1.0\textwidth]{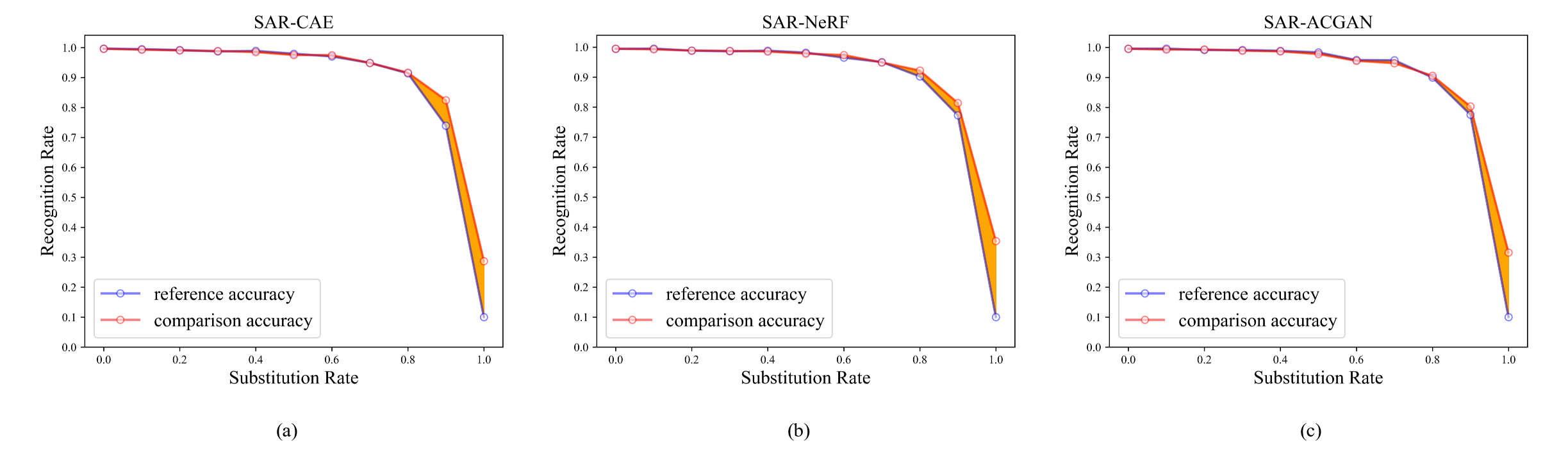}
    \caption{We applied the recognition versus substitution rate curves (RSC) \cite{guo2022recognition} to evaluate the simulated SAR image datasets in terms of utility, i.e., (a) SAR-NeRF, (b) SAR-CAE, and (c) SAR-ACGAN. The larger area between the reference accuracy curve and the comparison accuracy curve denotes the better utility of simulated data.}
    \label{fig:RSC}
\end{figure*}

Some researches also proposed the evaluation metrics based on data utility, such as the recognition versus substitution rate curves (RSC) \cite{guo2022recognition}. Fig. \ref{fig:RSC} illustrates the RSC results of three simulated SAR image datasets. Note that RSC is a dataset utility evaluation index which can only report the global evaluation result, rather than assigning criteria to individuals. To this end, it cannot filter out samples with low-utility. On the contrary, our method can evaluate data utility for each sample with a quantitative metric so as to filter the high quality data.


\begin{figure*}[!htbp]
    \centering
    \includegraphics[width=0.9\textwidth]{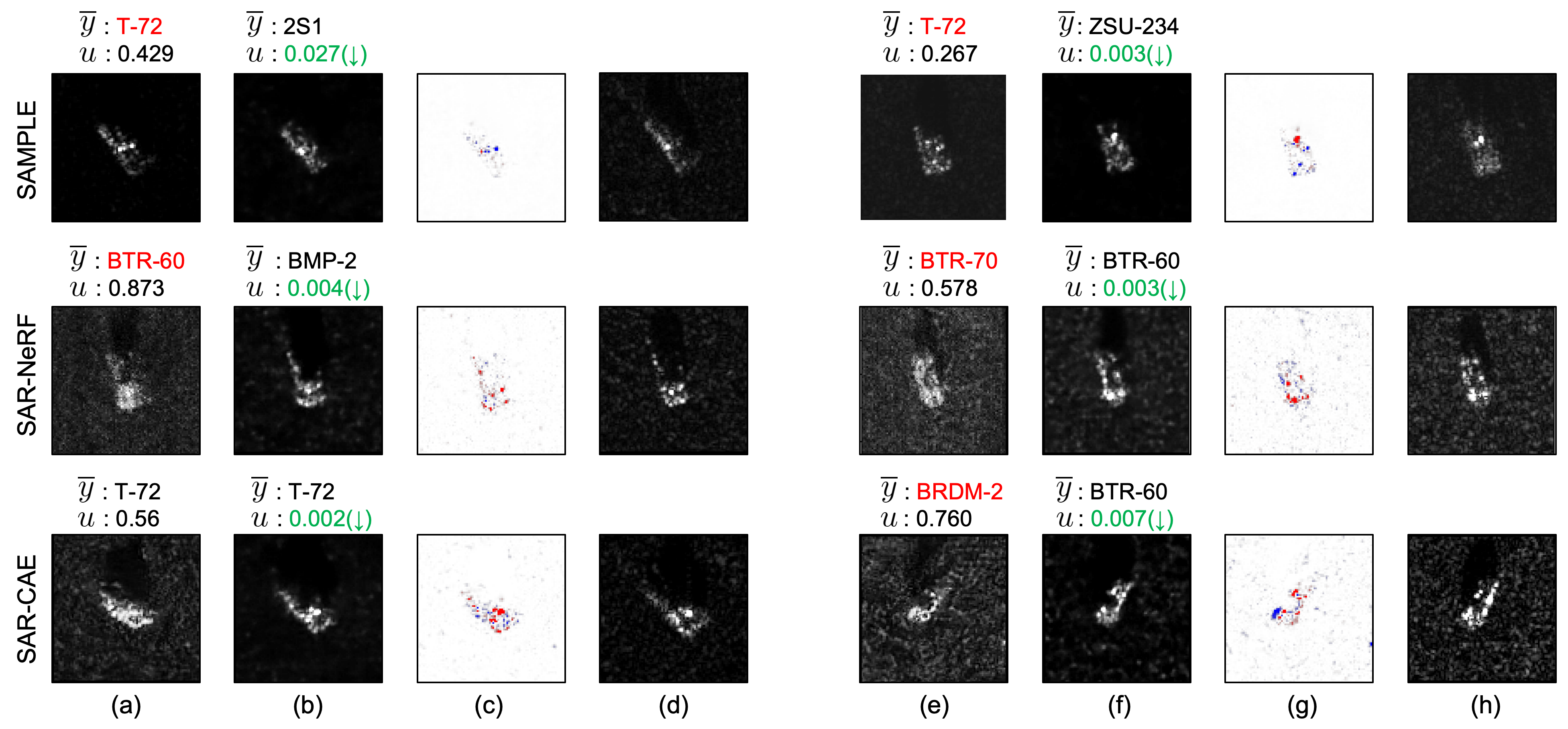}
    \caption{(a) and (e): Original simulated samples. (b) and (f): Generated counterfactual samples. (c) and (g): Inauthentic details. (d) and (h): Real samples. \textcolor{red}{Red}: wrong prediction. \textcolor{green}{Green}: decreased uncertainty.}
    \label{fig:Counterfactuals}
\end{figure*}

\subsection{Qualitative Results of Explanation}

To demonstrate that the proposed explanation method can explicitly reveal the inauthentic details of the simulated data, we visualize some examples from different simulated datasets, as shown in Fig. \ref{fig:Counterfactuals}. In this figure, we present the original simulated images in (a) and (e); generate counterfactual samples in (b) and (f); explain the inauthentic details in (c) and (g), where red and blue represent the missing and redundant parts of the simulated images, respectively; and present real samples with the closest azimuth angles in (d) and (h). The predicted uncertainty $u$ and class label $\overline{y}$ of the images are also provided.

It can be found that the uncertainty of the generated counterfactual image is reduced to a certain extent (marked in green). Additionally, the wrong prediction can be corrected (marked in red). Compared with the original simulated image, the counterfactual explanation is closer to the real one from a visual perspective. The inauthentic details explicitly explain the reason for the low utility of simulated data.

\begin{figure*}[!hbp]
    \centering
    \includegraphics[width=0.77\textwidth]{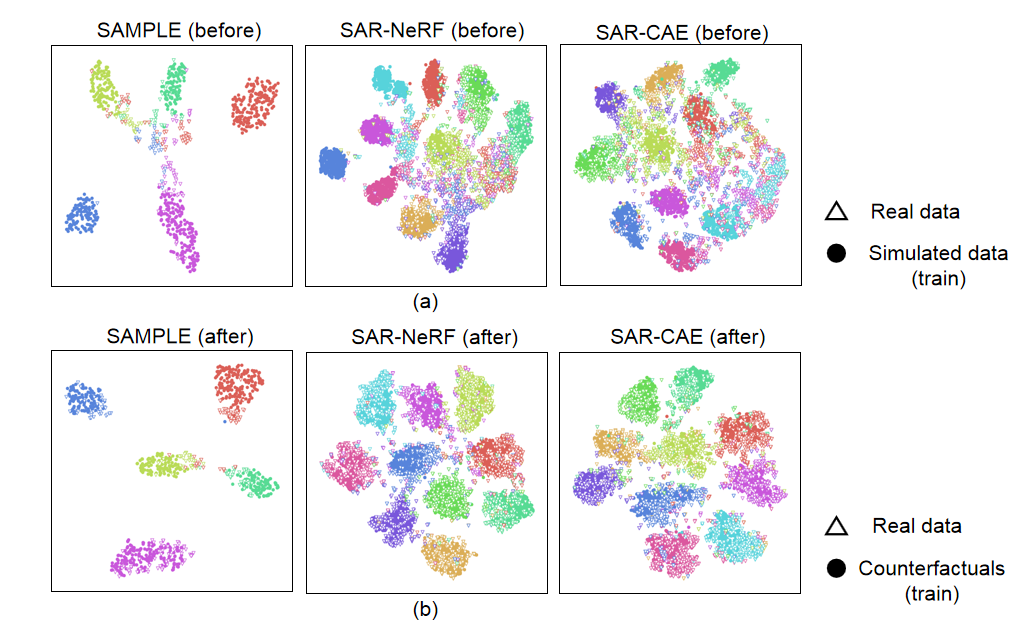}
    \caption{(a) The feature visualization of the real data and the original simulated samples demonstrates the inconsistent feature distribution between real and simulated SAR images. (b) The feature visualization of the real data and the counterfactual samples of simulated data indicates that the feature distribution becomes more consistent.}
    \label{fig:feature}
\end{figure*}

In addition, we verify the effectiveness of the proposed counterfactual explanation method by visualizing the feature distribution in the latent space. As provided in Fig. \ref{fig:feature}, the feature distribution of real and original simulated SAR images, and the one of real and counterfactual explanations, are demonstrated in (a) and (b), respectively. It can be seen that the original simulated data and the real ones have a distinct feature distribution inconsistency in the latent space. In contrast, the generated counterfactual samples achieve a more consistent distribution with the real ones, thus improving the data utility.

As illustrated above, the generated counterfactual explanation can improve the data quality and utility both explicitly and implicitly. Accordingly, we conducted several quantitative experiments to verify the potential of our proposed explanation method for improving the simulated data. In Table \ref{tab:counterfactual1}, we applied the A-ConvNet model trained on different simulated data to test the real MSTAR target images (15$^\circ$) and recorded the results. For each simulated dataset, four experiments are conducted. The term \textit{Upperbound} denotes training with full real data, which demonstrates the recognition model's upper limit. \textit{Before} denotes training with full original simulated SAR images, while \textit{after (BBB/MCD)} denotes training with counterfactual samples generated by Eva-BBB and Eva-MCD, respectively. We maintain the same number of training samples across all four experiments to ensure fairness in comparison. We can see that training with full original simulated data will cause a dramatic decrease in classification accuracy tested on real SAR images. The generated counterfactual explanation, however, can successfully improve the data utility that makes the trained model more applicable.

Furthermore, we also tested the performance of training different deep learning architectures with counterfactual samples, as given in Table \ref{tab:counterfactual2}. The experimented simulated dataset is SAR-NeRF. We explore popular convolutional neural network and vision transformer models, such as ResNet-18, ResNet-50, DenseNet-121, and Swin-Transformer, for illustration. The results also demonstrate the effectiveness of the proposed explanation method, which has the potential to improve the utility of the data.


\begin{table*}[!htbp]
\centering
\caption{The recognition rate of real data tested on A-ConvNet models with different training data. UpperBound: model trained with real data; Before: model trained with original simulated data; After: model trained with counterfactual samples. BBB/MCD denote using Eva-BBB and Eva-MCD as the evaluator respectively. \textcolor{red}{RED}: the best overall performance.}
\label{tab:counterfactual1}
\resizebox{\linewidth}{!}{
\begin{tabular}{ccccccccccccc}
\toprule
\multicolumn{2}{c}{Dataset}                                 & 2S1         & BMP-2       & BRDM-2     & BTR-60     & BTR-70      & D7         & T-62       & T-72        & ZIL-131     & ZSU-234     & Overall Acc \\
\midrule
\multicolumn{1}{r}{\multirow{4}{*}{SAR-NeRF}} &\cellcolor{gray!20}UpperBound  & \cellcolor{gray!20}90.27±1.05  & \cellcolor{gray!20}98.29±0.64  & \cellcolor{gray!20}94.77±2.11 & \cellcolor{gray!20}98.97±0.00 & \cellcolor{gray!20}98.98±0.83  & \cellcolor{gray!20}98.54±0.79 & \cellcolor{gray!20}97.92±1.25 & \cellcolor{gray!20}99.32±0.64  & \cellcolor{gray!20}100.00±0.00 & \cellcolor{gray!20}100.00±0.00 & \cellcolor{gray!20}97.55±0.30  \\
                         & before      & 44.89±6.03  & 73.33±13.18 & 76.40±3.49 & 90.09±2.79 & 91.16±3.01  & 39.54±5.83 & 3.54±0.91  & 85.37±5.84  & 9.49±2.60   & 65.45±11.65 & 54.45±0.77  \\
                        & after (BBB) & 89.66±0.17  & 92.82±2.55  & 85.89±3.78 & 93.85±1.45 & 94.73±0.96  & 87.96±0.60 & 66.30±1.30 & 96.94±0.72  & 49.03±0.91  & 90.88±5.20  & 83.55±1.04  \\
                       & after (MCD) & 94.77±1.75  & 91.97±3.41  & 90.02±1.34 & 92.14±2.07 & 95.92±2.73  & 90.27±4.18 & 71.31±5.43 & 96.43±1.25  & 55.11±10.34 & 95.01±2.26  & \textcolor{red}{86.42±0.96}  \\
\midrule
\multirow{4}{*}{SAR-CAE}                          & \cellcolor{gray!20}UpperBound  & \cellcolor{gray!20}94.53±1.37  & \cellcolor{gray!20}97.61±1.05  & \cellcolor{gray!20}96.35±2.06 & \cellcolor{gray!20}98.20±1.42 & \cellcolor{gray!20}99.17±0.00  & \cellcolor{gray!20}98.91±0.30 & \cellcolor{gray!20}98.29±0.91 & \cellcolor{gray!20}99.66±0.48  & \cellcolor{gray!20}99.88±0.17  & \cellcolor{gray!20}100.00±0.00 & \cellcolor{gray!20}98.18±0.22  \\
                                              & before      & 62.77±3.17  & 47.18±7.89  & 48.91±12.07 & 67.18±4.61 & 34.69±10.09   & 97.45±3.15 & 12.09±7.00 & 32.14±9.64  & 25.18±10.06  & 42.70±7.53 & 47.26±3.62  \\
                                              & after (BBB) & 90.80±3.13  & 73.74±5.70  & 69.20±8.89 & 77.44±2.53 & 76.02±4.48 & 96.50±2.01 & 81.61±2.91 & 80.51±3.31  & 83.50.29±2.63  & 91.02±3.92  & 82.70±1.16  \\
                                              & after (MCD) & 91.61±1.77  & 80.72±10.09  & 75.40±9.47 & 78.36±4.09 & 74.90±8.00 & 96.35±1.71 & 86.30±1.79 & 78.37±9.90  & 91.90±2.04  & 97.08±1.34  & \textcolor{red}{86.00±2.83}  \\
\midrule
\multirow{4}{*}{SAR-ACGAN}                        & \cellcolor{gray!20}UpperBound  & \cellcolor{gray!20}88.32±3.51  & \cellcolor{gray!20}97.44±0.73  & \cellcolor{gray!20}91.00±0.46 & \cellcolor{gray!20}97.78±0.64 & \cellcolor{gray!20}98.47±0.42  & \cellcolor{gray!20}98.66±0.46 & \cellcolor{gray!20}94.87±2.85 & \cellcolor{gray!20}98.13±1.46  & \cellcolor{gray!20}97.57±0.96  & \cellcolor{gray!20}98.66±0.34  & \cellcolor{gray!20}95.85±0.90  \\
                                              & before      & 38.98±5.95  & 19.38±5.47  & 69.20±5.97 & 54.87±4.71 & 30.82±4.70  & 93.57±1.71 & 12.09±7.74 & 44.29±8.56  & 52.41±8.16  & 52.41±10.00  & 48.04±0.36  \\
                                              & after (BBB) & 47.74±4.08  & 93.03±1.64  & 88.83±1.69 & 78.97±4.41 & 93.98±1.69  & 95.99±1.89 & 65.93±7.06 & 53.57±4.63  & 55.11±6.65  & 79.27±4.95  & 74.64±0.19  \\
                                              & after (MCD)  & 51.61±5.02  & 89.64±2.23  & 91.97±0.83 & 75.28±3.72 & 91.22±2.22  & 95.99±1.44 & 62.78±8.40 & 58.98±5.14  & 59.93±7.54  & 75.55±4.57  &\textcolor{red}{74.85±0.18}  \\
\midrule 
\multirow{4}{*}{SAMPLE}                       & \cellcolor{gray!20}UpperBound  & \cellcolor{gray!20}93.94±1.71  & \cellcolor{gray!20}100.00±0.00 &\cellcolor{gray!20} -          & \cellcolor{gray!20}-          & \cellcolor{gray!20}100.00±0.00 & \cellcolor{gray!20}-          & \cellcolor{gray!20}-          &\cellcolor{gray!20} 100.00±0.00 & \cellcolor{gray!20}-           & \cellcolor{gray!20}100.00±0.00 & \cellcolor{gray!20}98.38±0.46  \\
                                             & before      & 57.58±14.65 & 26.88±11.88 & -          & -          & 56.86±25.45 & -          & -          & 16.67±11.51 & -           & 92.59±6.05  & 55.66±3.37  \\
                                              & after (BBB) & 56.36±13.19 & 98.92±1.52  & -          & -          & 94.12±2.40  & -          & -          & 84.38±9.20  & -           & 100.00±0.00 & 84.79±5.10  \\
                                              & after (MCD) & 69.70±10.43 & 79.57±4.02  & -          & -          & 100.00±0.00 & -          & -          & 87.50±2.55  & -           & 96.30±5.24  & \textcolor{red}{85.92±3.15} \\
\bottomrule 
\end{tabular}}
\end{table*}

\begin{table*}[!htbp]
\centering
\caption{The recognition rate of real data tested on different deep models. UpperBound: model trained with real data; Before: model trained with original SAR-NeRF dataset; After: model trained with counterfactual samples of SAR-NeRF. \textcolor{red}{RED}: the best overall performance.}
\label{tab:counterfactual2}
\resizebox{\linewidth}{!}{
\begin{tabular}{clccccccccccc}
\toprule
\multicolumn{2}{c}{Model}                  & 2S1         & BMP-2       & BRDM-2      & BTR-60      & BTR-70      & D7          & T-62        & T-72        & ZIL-131    & ZSU-234     & Overall Acc \\
\midrule
\multirow{3}{*}{ResNet-18}    & \cellcolor{gray!20}UpperBound & \cellcolor{gray!20}78.95±5.67  & \cellcolor{gray!20}92.65±0.24  & \cellcolor{gray!20}95.50±1.75  & \cellcolor{gray!20}96.75±2.38  & \cellcolor{gray!20}100.00±0.00 & \cellcolor{gray!20}83.70±4.00  & \cellcolor{gray!20}90.48±2.26  & \cellcolor{gray!20}99.32±0.24  & \cellcolor{gray!20}93.55±2.82 & \cellcolor{gray!20}100.00±0.00 & \cellcolor{gray!20}92.56±0.64  \\
                              & before     & 5.47±5.95   & 36.92±3.65  & 70.19±19.15 & 71.62±5.87  & 98.47±0.42  & 22.87±5.85  & 0.00±0.00   & 92.52±2.84  & 0.49±0.34  & 77.86±5.86  & 44.15±1.15  \\
                              & after      & 81.51±7.49  & 79.49±2.75  & 79.93±10.58 & 87.69±1.51  & 94.56±3.76  & 80.41±7.94  & 58.36±12.68 & 94.22±2.29  & 44.89±6.11 & 98.66±0.62  & \textcolor{red}{78.82±1.37}  \\
\midrule
\multirow{3}{*}{ResNet-50}    & \cellcolor{gray!20}UpperBound & \cellcolor{gray!20}61.68±15.75 &\cellcolor{gray!20}87.52±5.46  & \cellcolor{gray!20}94.16±3.87  & \cellcolor{gray!20}98.80±0.64  &\cellcolor{gray!20}98.64±1.27  & \cellcolor{gray!20}78.83±8.82  & \cellcolor{gray!20}64.71±13.11 & \cellcolor{gray!20}97.79±1.05  & \cellcolor{gray!20}91.97±5.82 & \cellcolor{gray!20}100.00±0.00 &\cellcolor{gray!20}86.35±4.41  \\
                              & before     & 2.68±2.09   & 44.10±14.72 & 34.55±12.08 & 52.14±23.75 & 97.96±2.20  & 8.88±6.90   & 0.00±0.00   & 82.65±13.01 & 0.00±0.00  & 49.15±19.09 & 33.10±3.51  \\
                              & after      & 76.28±3.28  & 81.03±8.36  & 70.56±12.22 & 91.62±4.30  & 88.78±4.02  & 75.67±3.75  & 53.24±11.01 & 93.03±1.34  & 23.84±4.84 & 92.82±0.75  & \textcolor{red}{72.89±1.80}  \\
\midrule
\multirow{3}{*}{DenseNet-121} & \cellcolor{gray!20}UpperBound & \cellcolor{gray!20}67.76±6.60  & \cellcolor{gray!20}82.05±4.66  & \cellcolor{gray!20}82.73±14.39 & \cellcolor{gray!20}96.07±1.35  & \cellcolor{gray!20}100.00±0.00 & \cellcolor{gray!20}81.02±9.12  & \cellcolor{gray!20}90.96±7.88  & \cellcolor{gray!20}97.79±1.46  & \cellcolor{gray!20}90.39±1.05 & \cellcolor{gray!20}99.88±0.17  & \cellcolor{gray!20}88.21±3.05  \\
                              & before     & 0.00±0.00   & 1.20±1.35   & 4.14±3.28   & 7.35±7.54   & 98.13±1.20  & 0.00±0.00   & 0.00±0.00   & 80.10±16.50 & 0.12±0.17  & 63.02±27.16 & 22.69±1.10  \\
                              & after      & 48.05±13.27 & 56.24±10.53 & 76.52±14.13 & 88.21±4.41  & 94.90±1.25  & 67.03±19.14 & 57.88±12.10 & 98.81±1.34  & 33.94±7.05 & 98.78±0.62  & \textcolor{red}{70.43±5.41}  \\
\midrule
\multirow{3}{*}{Swin-T}       & \cellcolor{gray!20}UpperBound & \cellcolor{gray!20}72.99±6.41  & \cellcolor{gray!20}71.79±14.56 & \cellcolor{gray!20}97.32±1.13  & \cellcolor{gray!20}98.63±0.64  & \cellcolor{gray!20}96.94±2.08  & \cellcolor{gray!20}92.58±3.56  & \cellcolor{gray!20}80.10±7.18  & \cellcolor{gray!20}97.96±1.50  & \cellcolor{gray!20}93.07±4.17 & \cellcolor{gray!20}100.00±0.00 & \cellcolor{gray!20}89.99±3.09  \\
                              & before     & 0.12±0.17   & 0.00±0.00   & 0.12±0.17   & 0.00±0.00   & 100.00±0.00 & 0.00±0.00   & 0.00±0.00   & 0.00±0.00   & 0.97±0.62  & 15.33±6.41  & 9.95±0.66   \\
                              & after      & 41.36±18.12 & 45.81±13.28 & 79.56±14.10 & 66.32±11.15 & 92.35±6.05  & 45.26±6.11  & 25.89±1.80  & 53.23±15.68 & 74.70±8.20 & 45.01±14.87 & \textcolor{red}{56.00±3.79} \\
\bottomrule
\end{tabular}}
\end{table*}

\begin{figure}[!htbp]
    \centering
    \includegraphics[width=0.5\textwidth]{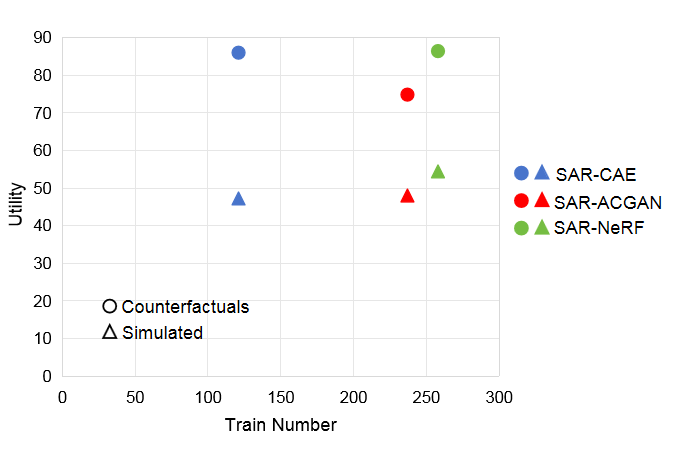}
    \caption{The x-axis denotes the amount of training data used to generate the simulated datasets. The y-axis denotes the data utility, which is the test results on real SAR images for a simulated data trained model.}
    \label{fig:data_v}
\end{figure}

Fig. \ref{fig:data_v} illustrates the utility evaluation results and the counterfactual explanation results of the three AI-generated SAR image datasets. The x-axis and the y-axis denote the number of training samples of the Gen-AI model, and the test results on real SAR images of a simulated data trained model, respectively. It can be observed that SAR-CAE data are generated by the fewest training samples and can perform comparable utility with SAR-ACGAN where the training data are doubled. With similar training numbers, SAR-NeRF data has better utility than SAR-ACGAN according to our evaluation results. The counterfactual explanations of SAR-NeRF and SAR-CAE data are more effective than those of SAR-ACGAN data, that improve the image utility better.

\subsection{Ablation Studies}

In order to verify the effectiveness of each module in the counterfactual generation process, we conducted a series of ablation experiments on the SAMPLE dataset, as shown in Table \ref{tab:Ablation Studies}.

First, the recognition model is trained directly on original simulated data and tested on real data to obtain the baseline result. Then, we applied the CLUE method to generate the counterfactual explanation, i.e., using VAE as the generator and optimizing with distance and uncertainty. The result shows a slight improvement of 5.15\% over the baseline, indicating the effectiveness of the counterfactual explanation primarily. It is worth noting that the utility of the generated counterfactual data has increased significantly after adding the class guidance. Additionally, IntroVAE can improve the visual quality of counterfactual samples compared with VAE, enhancing its utility as well as obtaining high-quality counterfactual interpretation samples.

By introducing the angle guidance, we need to enable the assessment model to predict the azimuth angle, denoted as Eva-BBB. It can be obviously observed from the fourth row and the last row that introducing angle guidance can further improve the utility of counterfactual samples. The performance of predicting class and angle has improved to a certain extent. Moreover, we also verify the superiority of BayesCNN based CNN in predicting uncertainty compared with point-estimated frequency neural networks, as given in the 5th and 6th rows in Table \ref{tab:Ablation Studies}.

\begin{table*}[!htbp]
\centering
\caption{Ablation studies.}
\label{tab:Ablation Studies}
\begin{tabular}{cccccc}
\toprule
\multicolumn{1}{l}{\textbf{IntroVAE}} & \multicolumn{1}{l}{\textbf{Class Guidance}} & \multicolumn{1}{l}{\textbf{Angle Guidance
}} & \textbf{P-Eva}    & \textbf{Accuracy}                                                      & \textbf{Angle loss}                                                      \\
\midrule
\multicolumn{3}{c}{Baseline (Train: Sim data, Test: Real data)}  & \XSolidBrush & 58.35 $\pm$ 4.11 &0.0086 $\pm$ 0.0019 \\
$\times$ & $\times$ & $\times$& Eva-BBB (w/o angle) & 63.50 $\pm$ 9.47 & 0.0107 $\pm$ 0.0008 \\$\times$  & \checkmark& $\times$& Eva-BBB (w/o angle) & 77.19 $\pm$ 4.79& 0.0115 $\pm$ 0.0018\\
\checkmark & \checkmark  & $\times$ & Eva-BBB (w/o angle) & 79.32 $\pm$ 2.31& 0.0090 $\pm$ 0.0015\\
\checkmark  & \checkmark& \checkmark  & CNN  & 77.86 $\pm$ 1.91 & 0.0095 $\pm$ 0.0009\\
\checkmark&\checkmark& \checkmark & Eva-BBB & \textbf{88.25 $\pm$ 1.44} & \textbf{0.0054 $\pm$ 0.0008}\\
\bottomrule
\end{tabular}
\end{table*}



\subsection{Hyper-Parameter Discussion}

There are three hyperparameters involved in the process of counterfactual sample generation. $\lambda_y$ and $\lambda_v$ control the trade-off between the classification loss and angle loss, and $\lambda_d$ controls the distance term between the counterfactual and original data. By fixing $\lambda_d=1$ and changing the ratio of $\lambda_y: \lambda_v$, it can be found that with the increase in the proportion of angle loss, the azimuth angle of the generated counterfactual image is more accurate, and the recognition model trained with these counterfactual samples performs better on azimuth angle regression prediction. When $\lambda_y:\lambda_v=1:30$, the result achieves the highest classification accuracy on the test set. We also experience the influence of $\lambda_d$. The result shows that a larger $\lambda_d$ would result in less change from the original image, leading to inferior counterfactual results.

\begin{table}[!htbp]
\centering
\caption{The hyper-parameter discussions.}
\label{tab:Hyper-parameter Discussion}
\begin{tabular}{cccc}
\toprule
\textbf{$\lambda_d$} & \textbf{$\lambda_y$ : $\lambda_v$} & \textbf{Accuracy}& \textbf{Angle loss}\\
\midrule
& 1:1 & 84.08$\pm$1.35 & 0.0059$\pm$0.0013 \\
& 1:30 & \textbf{88.25$\pm$1.44} & 0.0054$\pm$0.0008 \\ 
\multirow{-3}{*}{$1$}    & 1:100 & 86.89$\pm$1.59 & \textbf{0.0051$\pm$0.0005} \\
\midrule
\multicolumn{1}{c}{$10$} & 1:30 & 85.44$\pm$2.42 & 0.0056$\pm$0.0014\\
\bottomrule
\end{tabular}
\end{table}





\section{Conclusion}
\label{sec:conclusion}

In this paper, a novel trustworthy framework, \textit{X-Fake}, is proposed to evaluate the simulated SAR images and explain the inauthentic details of them from the perspective of data utility. The proposed framework comprises a Bayesian convolutional neural network-based probabilistic evaluator to output predicted uncertainty and a generative model-based causal explainer to obtain the counterfactual explanation. The predicted uncertainty of simulated data reveals an inconsistent distribution with real ones, which can be regarded as a criteria for utility assessment. The counterfactual explanations are generated with prior information, including the target label, azimuth angle, and the predicted uncertainty from the evaluator. It demonstrates the inauthentic details of the simulated image that cause the inconsistent data distribution. Experiments are conducted on several simulated SAR image datasets generated by both EM-based and GenAI-based approaches. The results illustrate that the proposed evaluator can successfully filter out high-quality data in terms of utility, while the trained deep learning model can achieve higher performance on real SAR images. Additionally, the generated counterfactual images explicitly improve the utility of simulated data. The classification model trained on generated counterfactual images outperforms the ones trained on original simulated data by 30\%.

\bibliography{IEEEabrv,submit}

\bibliographystyle{IEEEtran}

\vfill

\end{document}